\documentclass[10pt,twocolumn,letterpaper]{article}

\usepackage{cvpr}              

\usepackage{amsmath}
\usepackage{amssymb}

\usepackage{graphicx}
\usepackage{booktabs}
\usepackage{multirow}
\usepackage{float}  
\usepackage{placeins} 
\usepackage{stfloats} 

\usepackage{pifont}

\usepackage{tikz}
\usetikzlibrary{arrows.meta,positioning,calc,fit,backgrounds,shapes.geometric,shapes.misc,decorations.pathreplacing}

\usepackage{enumitem}

\DeclareMathOperator{\CrossAttn}{CrossAttn}

\definecolor{cvprblue}{rgb}{0.21,0.49,0.74}
\usepackage[pagebackref,breaklinks,colorlinks,allcolors=cvprblue]{hyperref}

\def\confName{CVPR}
\def\confYear{2026}

\title{Hand Trajectory Fusion for Egocentric Natural Language Query Grounding}

\author{ 
    Enmin Zhong, Carlos R. del-Blanco, Fernando Jaureguizar, Narciso García \\
    Grupo de Tratamiento de Imágenes (GTI), Information Processing and Telecommunications Center , \\
    ETSI Telecomunicación, Universidad Politécnica de Madrid, Spain \\
    \tt\small \{enmin.zhong, carlosrob.delblanco, fernando.jaureguizar, narciso.garcia\}@upm.es
}

\begin{document}
\maketitle

\begin{abstract}

Egocentric Natural Language Query (NLQ) grounding asks a model to localize, in a long first-person video, the temporal interval that answers a free-form text query. 
Existing methods fuse video appearance with the query but ignore hand motion, despite the fact that roughly $41\%$ of Ego4D NLQ queries are answered at a moment of hand--object manipulation or their immediate outcomes.
We propose a hand-trajectory encoder for converting a sequence of
hand skeletons into highly-semantic hand kinematic features, which
are then aligned and combined with pretrained video--text features
through a cross-attention fusion strategy with adaptive gating.
On the Ego4D NLQ v2 validation split, the clearest gains appear
for Hand-Object Interaction queries ($+2.54$ R1@IoU=0.3) and
Quantity/State queries ($+4.32$ R1@IoU=0.3), indicating that hand
trajectory provides grounding cues beyond appearance alone.

\end{abstract}



\section{Introduction}
\label{sec:intro}

First-person video records the world from the perspective of the hands.
When a person searches their memory for ``What did I put in the box?'' or ``Where is the red screwdriver?'', the answer is grounded in a specific moment of manual activity --- reaching, grasping, and placing.
Natural Language Query (NLQ) grounding on Ego4D~\cite{Grauman2022} formalizes this problem: given a text query and a long egocentric video clip, the model must predict the answer span $[t_s, t_e]$ where the queried activity occurred.

State-of-the-art NLQ systems such as GroundNLQ~\cite{GroundNLQ} rely on large pretrained video encoders (InternVideo~\cite{Wang2022InternVideo}, EgoVLP~\cite{Lin2022EgoVLP}) fused with CLIP text features.
These models excel at matching semantic appearance but lack explicit access to auxiliary modalities that are meaningful for many queries.
Recent works address this gap by injecting dense or spatially grounded signals: GazeNLQ~\cite{GazeNLQ} adds predicted gaze information to video-text features via a dedicated encoder and then uses residual cross-attention for information fusion; ObjectNLQ~\cite{Feng2024ObjectNLQ} introduces an object-detection branch that combines and encodes object detections in frames obtained by a Co-DETR\cite{codet} detector with CLIP-based text features, so that query-relevant object information is emphasized; lastly OSGNet~\cite{OSGNet} inherits this motivation but extends it with an additional shot branch that models egocentric camera/head movement as a proxy for wearer attention. However, no published work has studied \emph{hand trajectory} as an auxiliary modality for NLQ grounding, despite hand motion being a primary cue in egocentric
activity. Although hand priors are well-established in adjacent egocentric tasks -- hand-object contact detection~\cite{Shan2020100DOH}, action
anticipation~\cite{Furnari2019RULSTM}, and kinematic pretraining~\cite{Pei2025EgoHOD} -- their application to \emph{temporal language grounding} remains unexplored.


\begin{figure}[t]
    \centering
    \includegraphics[width=\linewidth]{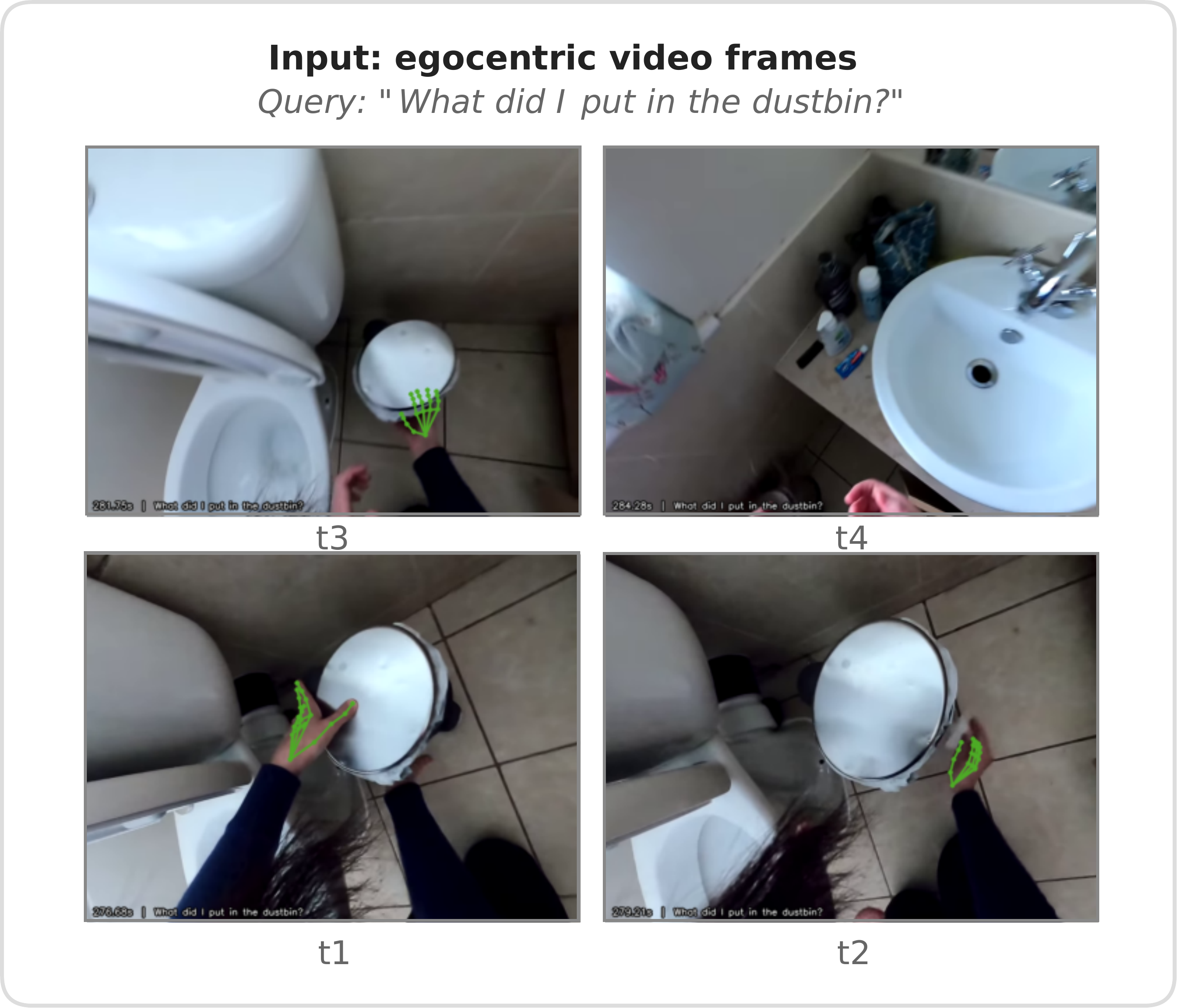}
    \caption{Hand trajectories across Hand-Object Interaction queries. The hand skeleton (green) provides a kinematic signal that is distinctive at the moment of manipulation and complementary to visual appearance. Notice also that hands are not detected in all frames.
    }
    \label{fig:frames_hoi}
\end{figure}



Indeed, among the 13 NLQ template types in Ego4D, five describe events whose ground-truth window is either a manipulation action or its immediate result (see Fig.~\ref{fig:frames_hoi}) ---``Where did I put X?'', ``What did I put in X?'', ``What X did I $\langle$action$\rangle$?'', ``What is the state of X?'', and ``Where is my object X?''. Together these five templates cover $7{,}529$ of the $18{,}315$ train+val queries, all answered at moments of hand--object contact. We refer to this union as \emph{manipulation-centric queries} throughout. 

However, it is a challenge to effectively use and combine hand information with visual and text ones. Existing hand skeleton extractors, such as Mediapipe~\cite{Zhang2020MediaPipe}, provide 21 anatomical landmarks per hand, but this information is sparse in time. On the Ego4D NLQ split, hands are detected in only 41\% of frames on average, due to long idle periods, motion blur, and out-of-frame hands.
In contrast to gaze (a dense 1-D scalar per frame) and object detections (multiple per-frame boxes), hand trajectory suffers from frequent gaps, complicating both the trajectory encoding and the fusion strategy with video--text information.


In this work, we address this challenge by adopting two design decisions:
(1) a trajectory encoder that models spatial relations among hand joints and temporal dynamics across frames in separate stages, while explicitly masking undetected frames; and (2) a fusion strategy that integrates trajectory features with the video--text representation through cross-attention and a learned gating mechanism, allowing the trajectory signal to contribute selectively to the prediction.

\section{Method}
\label{sec:method}

\begin{figure*}[!t]
    \centering
    \includegraphics[width=\linewidth]{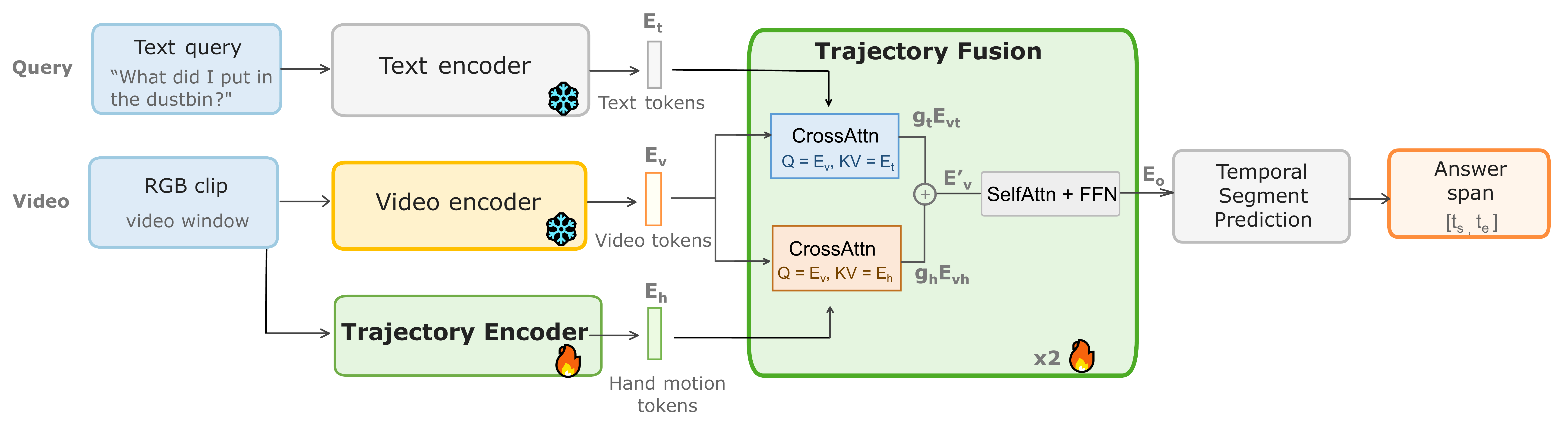}
    \caption{Overall architecture of the proposed hand-trajectory NLQ grounding model. $g_h$ and $g_t$ denote the learned scalar gates from Eq.~\eqref{eq:gated-residual}.}
    \label{fig:architecture}
\end{figure*}

The proposed approach grounds natural-language queries in
egocentric video by combining hand trajectory with video-text
semantic features, so that queries related to object manipulation
can be temporally localized more accurately.
\Cref{fig:architecture} illustrates the system, organized in five
modules. The \emph{Video Encoder} embeds the egocentric clip into a
sequence of video tokens $\mathbf{E_v}$ using the pretrained and
frozen InternVideo~\cite{Wang2022InternVideo} and
EgoVLP~\cite{Lin2022EgoVLP} backbones, and the \emph{Text Encoder}
embeds the natural-language query into text tokens $\mathbf{E_t}$
using the pretrained and frozen CLIP~\cite{Radford2021CLIP}. In
parallel, the trainable \emph{Trajectory Encoder}
(\cref{sec:traj_enc}) takes the temporal sequence of hand
skeletons~\cite{Zhang2020MediaPipe} and produces video-aligned
kinematic features $\mathbf{E_h}$. The trainable \emph{Trajectory
Fusion} module (\cref{sec:fusion}) integrates $\mathbf{E_h}$ and
$\mathbf{E_t}$ into $\mathbf{E_v}$ through cross-attention with
adaptive gating, followed by a self-attention refinement, yielding
the multimodal representation $\mathbf{E_o}$. Finally, the
\emph{Temporal Segment Prediction} module predicts from $\mathbf{E_o}$ the answer span $[t_s, t_e]$ that best matches the query.

\subsection{Trajectory Encoder}
\label{sec:traj_enc}
The Trajectory Encoder converts the sparse sequence of hand skeletons produced by the Hand Skeleton Extractor~\cite{Zhang2020MediaPipe} into a dense, video-aligned kinematic representation $\mathbf{E}_h \in \mathbb{R}^{D \times T}$, where $T$ is the number of video frames and $D$ is the latent dimension shared with the rest of the architecture.
We adopt a spatio-temporal transformer that factorizes the problem into two stages: spatial cross-attention aggregates the landmarks of each frame into a single descriptor, and temporal self-attention then models how this descriptor evolves across frames.
This factorization mirrors the structure of manipulation events, whose semantics arise from how a static hand configuration changes over time —approach, contact, release.

\paragraph{Input and joint tokenization.}
For each frame $t \in \{1,\dots,T\}$, the encoder receives up to $L = 2 \times 21 = 42$ landmarks, indexed by $\ell \in \{1,\dots,L\}$ so that each value of $\ell$ uniquely identifies a (hand, joint) pair. Each landmark is described by its raw channels $\mathbf{r}_{t,\ell} = (x,y,z,v) \in \mathbb{R}^{4}$, encoding 3D location and visibility, and is embedded into a $D$-dimensional token as
\begin{equation}
\mathbf{x}_{t,\ell}
= \mathbf{W}_{r,\ell}\,\mathbf{r}_{t,\ell}
+ \mathbf{p}_{\ell},
\label{eq:joint-token}
\end{equation}
where $\mathbf{W}_{r,\ell} \in \mathbb{R}^{D \times 4}$ is a per-landmark learnable projection that jointly encodes the raw kinematic channels together with the identity of the corresponding (hand, joint) pair, and $\mathbf{p}_{\ell} \in \mathbb{R}^{D}$ is a positional encoding that disambiguates landmarks in the spatial attention that follows.

\paragraph{Spatial aggregation.}
A shared learnable query $\mathbf{q} \in \mathbb{R}^{D}$ pools the $L$ landmark tokens of each frame via cross-attention,
\begin{equation}
\mathbf{s}_t = \mathrm{CrossAttn}\!\left(
\mathbf{Q}{=}\mathbf{q},\,
\mathbf{K}{=}\mathbf{V}{=}\{\mathbf{x}_{t,\ell}\}_{\ell=1}^{L}
\right) \in \mathbb{R}^{D},
\label{eq:spatial-pool}
\end{equation}
yielding a frame-level descriptor that emphasizes the most informative joints (e.g., fingertips during a grasp) instead of committing to a fixed pooling rule. Undetected landmarks are excluded through a key-padding mask.

\paragraph{Temporal modeling.}
The descriptors $\{\mathbf{s}_t\}_{t=1}^{T}$ are then refined by a temporal self-attention layer and linearly projected to the kinematic features $\mathbf{E}_h \in \mathbb{R}^{D \times T}$,
\begin{equation}
\mathbf{E}_h
= \mathrm{Proj}\!\left(
\mathrm{SelfAttn}\!\left(\{\mathbf{s}_t\}_{t=1}^{T}\right)
\right),
\label{eq:temporal-attn}
\end{equation}
capturing the multi-frame structure of manipulation events. An analogous mask prevents frames with no detected hand from leaking into the temporal context.

\subsection{Trajectory Fusion}
\label{sec:fusion}

The Trajectory Fusion module injects the kinematic context
$\mathbf{E_h}$ and the text query $\mathbf{E_t}$ into the video
tokens $\mathbf{E_v}$, producing a multimodal representation
$\mathbf{E_o}$. Its design is driven by two requirements:
preserving the video--text alignment that the prediction head
relies on, and letting the model learn how strongly to rely on the
trajectory branch depending on the clip content. We address both
by querying the auxiliary modalities from $\mathbf{E_v}$ via
cross-attention, and modulating their contribution with two
learned, content-dependent gates. The block is stacked twice, and
the final output is fed to the Temporal Segment Prediction head.

\paragraph{Cross-attention and adaptive gating.}
Two cross-attention modules let the video tokens query the
trajectory and text streams independently,
\begin{equation}
\begin{split}
\mathbf{E_{vh}} &= \CrossAttn(\mathbf{Q}{=}\mathbf{E_v},\,
                              \mathbf{K}{=}\mathbf{V}{=}\mathbf{E_h}),\\
\mathbf{E_{vt}} &= \CrossAttn(\mathbf{Q}{=}\mathbf{E_v},\,
                              \mathbf{K}{=}\mathbf{V}{=}\mathbf{E_t}),
\end{split}
\label{eq:cross-attn}
\end{equation}
yielding two video-aligned representations enriched with kinematic
and semantic context. The two outputs are then added to
$\mathbf{E_v}$ through a residual connection in which the
contribution of each branch is scaled by a learned gate, rather
than summed uniformly. Each gate is produced by a lightweight MLP
applied to the temporally averaged output of its own
cross-attention,
$g_h = \sigma(\text{MLP}_h(\bar{\mathbf{e}}_{vh}))$ and
$g_t = \sigma(\text{MLP}_t(\bar{\mathbf{e}}_{vt}))$, where
$\bar{\mathbf{e}}_{vh}, \bar{\mathbf{e}}_{vt} \in \mathbb{R}^{D}$
are the temporal averages of $\mathbf{E_{vh}}$ and $\mathbf{E_{vt}}$,
$\sigma$ is the sigmoid, and $g_h, g_t \in (0,1)$. The merged
representation is
\begin{equation}
\mathbf{E_v}'
= \mathbf{E_v}
+ g_h \cdot \mathbf{E_{vh}}
+ g_t \cdot \mathbf{E_{vt}}.
\label{eq:gated-residual}
\end{equation}
Because each gate reads its own branch, the network can attenuate
one branch independently of the other ---e.g., when hands are
mostly undetected and $\mathbf{E_{vh}}$ carries little signal.

\paragraph{Self-attention refinement.}
A standard transformer block, $f_{\text{self}}(\cdot)$, refines the
merged representation through self-attention and a feed-forward
network,
\begin{equation}
\mathbf{E_o} = \mathbf{E_v}' + f_{\text{self}}(\mathbf{E_v}'),
\label{eq:fusion-self}
\end{equation}
yielding the fused multimodal representation $\mathbf{E_o}$.

\paragraph{Training.}
The Trajectory Encoder comprises 195K parameters
(0.6\% of the full model) and is trained from scratch jointly
with the Trajectory Fusion module using AdamW
($\text{lr} = 5 \times 10^{-5}$, cosine decay, 2 warmup epochs, with a $2\times$ higher learning rate for newly introduced modules).

\section{Experiments}
\label{sec:experiments}

The proposed system is evaluated on Ego4D NLQ v2~\cite{Grauman2022}, which contains 13{,}435 train and 4{,}552 validation query-clip pairs; training is performed on the training split, and results are reported on the validation split. The used metric is the standard R$m$@IoU=$n$: the percentage of queries for which at least one of the top-$m$ predicted moments has IoU $\geq n$ with the ground truth, evaluated at thresholds $n=0.3$ and $n=0.5$. 
 
To test the central hypothesis that hand kinematics help action-centric grounding, per-category R1 is reported on the two Ego4D categories closest to the manipulation-centric query set defined in Sec.~\ref{sec:intro}: \textbf{Hand--Object Interaction} (HOI; $N{=}1{,}928$), describing what the camera wearer did with an object, and \textbf{Quantity/State} ($N{=}718$), describing object counts or states. 

\Cref{tab:category} compares the GroundNLQ baseline---reproduced
locally without the trajectory branch---against the proposed
hand-trajectory model. The largest gains appear precisely in these
categories: $+2.54$ R1@IoU=0.3 on HOI and $+4.32$ on Quantity/State, consistent with kinematics encoding the approach--grasp--release pattern that is temporally distinctive at the moment of contact. Within HOI, the gain is concentrated in action templates (\emph{``What X did I $\langle$action$\rangle$?''}: $+4.00$; \emph{``What did I put in X?''}: $+4.58$), confirming that trajectory primarily helps localize when an action happened.

\Cref{tab:overall} reports the overall comparison with the
GroundNLQ baseline. Beyond the per-category gains, the proposed
model improves R1@0.5 by $+1.39$, almost twice the gain at
R1@0.3 ($+0.77$), indicating that hand kinematics not only help
retrieve the relevant temporal region but also sharpen the
localization at the moment of manipulation.

\begin{table}[t]
\centering
\caption{Per-category R1 on Ego4D NLQ v2 validation split.}
\label{tab:category}
\small
\setlength{\tabcolsep}{4pt}
\renewcommand{\arraystretch}{1.1}
\resizebox{\columnwidth}{!}{%
\begin{tabular}{@{}lrcccccc@{}}
\toprule
\textbf{Category} & $N$ &
\multicolumn{3}{c}{\textbf{R1@0.3}} &
\multicolumn{3}{c}{\textbf{R1@0.5}} \\
\cmidrule(lr){3-5} \cmidrule(lr){6-8}
& & GroundNLQ & Ours & $\Delta$ & GroundNLQ & Ours & $\Delta$ \\
\midrule
HOI               & 1928 & 28.99 & 31.54 & $+$2.54 & 19.97 & 21.73 & $+$1.76 \\
Quantity / State  &  718 & 24.93 & 29.25 & $+$4.32 & 16.85 & 21.17 & $+$4.32 \\
\midrule
\textbf{Overall}  & 4552 & 25.77 & 26.54 & $+$0.77 & 17.11 & 18.50 & $+$1.39 \\
\bottomrule
\end{tabular}%
}
\end{table}


\begin{table}[t]
\centering
\caption{Overall comparison with GroundNLQ on Ego4D NLQ v2
validation split.}
\label{tab:overall}
\small
\setlength{\tabcolsep}{6pt}
\renewcommand{\arraystretch}{1.1}
\begin{tabular}{@{}lccc@{}}
\toprule
\textbf{Model} & \textbf{R1@0.3} & \textbf{R1@0.5} & \textbf{R5@0.3} \\
\midrule
GroundNLQ (baseline) & 25.77 & 17.11 & 51.87 \\
Ours (traj.\ fusion) & \textbf{26.54} & \textbf{18.50} & \textbf{52.37} \\
\bottomrule
\end{tabular}
\end{table}

\section{Conclusion}
\label{sec:conclusion}
Hand kinematics provide a lightweight yet informative signal for
egocentric NLQ grounding. A 195K-parameter Trajectory Encoder maps
raw hand landmarks into video-aligned kinematic features, and a
Trajectory Fusion module integrates them with video and query
tokens through cross-attention and adaptive gating, while leaving
the pretrained backbone frozen. On Ego4D NLQ v2, this design yields its largest gains exactly where the prior predicts they should appear: $+2.54$ R1@IoU=0.3 on Hand--Object Interaction queries and $+4.32$ on Quantity/State, jointly covering ${\approx}\,41\%$ of the validation set.

The main limitation is detection sparsity: hands are visible in
only $41\%$ of frames, capping how much the trajectory branch can
contribute. Improvements in egocentric hand detection should
translate directly into stronger grounding. Beyond this, the
Trajectory Fusion module is modality-agnostic and extends naturally to complementary signals such as gaze, as well as to larger-scale training, without modifying the pretrained backbone.

\section*{Funding}
This work was supported in part by Ministerio de Ciencia, Innovaci\'{o}n y Universidades (MCIU)/Agencia Estatal de Investigaci\'{o}n of the Spanish Government under projects PID2023-148922OA-I00 (EEVOCATIONS) and by Comunidad de Madrid under project TEC-2024/COM-322 (IDEALCV-CM).


{
    \small
    \bibliographystyle{ieeenat_fullname}
    \bibliography{main}

@inproceedings{Grauman2022,
  title={Ego4d: Around the world in 3,000 hours of egocentric video},
  author={Grauman, Kristen and Westbury, Andrew and Byrne, Eugene and Chavis, Zachary and Furnari, Antonino and Girdhar, Rohit and Hamburger, Jackson and Jiang, Hao and Liu, Miao and Liu, Xingyu and others},
  booktitle={Proceedings of the IEEE/CVF conference on computer vision and pattern recognition},
  pages={18995--19012},
  year={2022}
}

@inproceedings{GroundNLQ,
  title     = {{GroundNLQ} @ Ego4D Natural Language Queries Challenge 2023},
  author    = {Hou, Zhifan and Luo, Lei and Yin, Da and others},
  booktitle = {CVPR Workshop on Egocentric Perception, Interaction and Computing (EPIC)},
  year      = {2023}
}

@article{Feng2024ObjectNLQ,
  title={Objectnlq@ ego4d episodic memory challenge 2024},
  author={Feng, Yisen and Zhang, Haoyu and Xie, Yuquan and Li, Zaijing and Liu, Meng and Nie, Liqiang},
  journal={arXiv preprint arXiv:2406.15778},
  year={2024}
}

@misc{GazeNLQ,
      title={GazeNLQ @ Ego4D Natural Language Queries Challenge 2025}, 
      author={Wei-Cheng Lin and Chih-Ming Lien and Chen Lo and Chia-Hung Yeh},
      year={2025},
      eprint={2506.05782},
      archivePrefix={arXiv},
      primaryClass={cs.CV},
}

@inproceedings{OSGNet,
  title={Object-shot enhanced grounding network for egocentric video},
  author={Feng, Yisen and Zhang, Haoyu and Liu, Meng and Guan, Weili and Nie, Liqiang},
  booktitle={Proceedings of the Computer Vision and Pattern Recognition Conference},
  pages={24190--24200},
  year={2025}
}

@inproceedings{codet,
  title={Detrs with collaborative hybrid assignments training},
  author={Zong, Zhuofan and Song, Guanglu and Liu, Yu},
  booktitle={Proceedings of the IEEE/CVF international conference on computer vision},
  pages={6748--6758},
  year={2023}
}

@article{Wang2022InternVideo,
  title={Internvideo: General video foundation models via generative and discriminative learning},
  author={Wang, Yi and Li, Kunchang and Li, Yizhuo and He, Yinan and Huang, Bingkun and Zhao, Zhiyu and Zhang, Hongjie and Xu, Jilan and Liu, Yi and Wang, Zun and others},
  journal={arXiv preprint arXiv:2212.03191},
  year={2022}
}

@article{Lin2022EgoVLP,
  title={Egocentric video-language pretraining},
  author={Lin, Kevin Qinghong and Wang, Jinpeng and Soldan, Mattia and Wray, Michael and Yan, Rui and Xu, Eric Z and Gao, Difei and Tu, Rong-Cheng and Zhao, Wenzhe and Kong, Weijie and others},
  journal={Advances in Neural Information Processing Systems},
  volume={35},
  pages={7575--7586},
  year={2022}
}

@inproceedings{Radford2021CLIP,
  title={Learning transferable visual models from natural language supervision},
  author={Radford, Alec and Kim, Jong Wook and Hallacy, Chris and Ramesh, Aditya and Goh, Gabriel and Agarwal, Sandhini and Sastry, Girish and Askell, Amanda and Mishkin, Pamela and Clark, Jack and others},
  booktitle={International conference on machine learning},
  pages={8748--8763},
  year={2021},
  organization={PmLR}
}

@inproceedings{Shan2020100DOH,
  title={Understanding human hands in contact at internet scale},
  author={Shan, Dandan and Geng, Jiaqi and Shu, Michelle and Fouhey, David F},
  booktitle={Proceedings of the IEEE/CVF conference on computer vision and pattern recognition},
  pages={9869--9878},
  year={2020}
}

@inproceedings{Furnari2019RULSTM,
  title={What would you expect? anticipating egocentric actions with rolling-unrolling lstms and modality attention},
  author={Furnari, Antonino and Farinella, Giovanni Maria},
  booktitle={Proceedings of the IEEE/CVF International conference on computer vision},
  pages={6252--6261},
  year={2019}
}

@article{Pei2025EgoHOD,
  title={Modeling fine-grained hand-object dynamics for egocentric video representation learning},
  author={Pei, Baoqi and Huang, Yifei and Xu, Jilan and Chen, Guo and He, Yuping and Yang, Lijin and Wang, Yali and Xie, Weidi and Qiao, Yu and Wu, Fei and others},
  journal={arXiv preprint arXiv:2503.00986},
  year={2025}
}

@article{Zhang2020MediaPipe,
  title={Mediapipe hands: On-device real-time hand tracking},
  author={Zhang, Fan and Bazarevsky, Valentin and Vakunov, Andrey and Tkachenka, Andrei and Sung, George and Chang, Chuo-Ling and Grundmann, Matthias},
  journal={arXiv preprint arXiv:2006.10214},
  year={2020}
}
}


\end{document}